\documentclass[11pt,a4paper]{article}
\usepackage{authblk}
\usepackage{hyperref}
\usepackage{acl2015}
\usepackage{times}
\usepackage{url}
\usepackage{graphicx}
\usepackage{caption}
\usepackage{latexsym}
\usepackage{amsmath}
\usepackage{multirow}

 \title {A POS Tagger for Code Mixed Indian Social Media Text \textemdash{} ICON-2016 NLP Tools Contest Entry from Surukam}

\author{Sree Harsha Ramesh}
\author{Raveena R Kumar}
\affil{Surukam Analytics, Chennai \protect\\

{\tt \{harsha,raveena\}@surukam.com}} 

\begin{document}
 \maketitle

\begin{abstract}
Building Part-of-Speech (POS) taggers for code-mixed Indian languages is a particularly challenging problem in computational linguistics due to a dearth of accurately annotated training corpora.\ ICON, as part of its NLP tools contest has organized this challenge as a shared task for the second consecutive year to improve the state-of-the-art. This paper describes the POS tagger built at Surukam to predict the coarse-grained and fine-grained POS tags for three language pairs \textemdash{} Bengali-English, Telugu-English and Hindi-English, with the text spanning three popular social media platforms \textemdash{} Facebook, WhatsApp and Twitter. We employed Conditional Random Fields as the sequence tagging algorithm and used a library called \textit{sklearn-crfsuite} \textemdash{} a thin wrapper around CRFsuite for training our model.
Among the features we used include \textemdash{} character n-grams, language information and patterns for emoji, number, punctuation and web-address. 
Our submissions in the constrained environment, i.e., without making any use of monolingual POS taggers or the like, obtained an overall average F1-score of 76.45\%, which is comparable to the 2015 winning score of 76.79\%. 
\end{abstract}

\section{Introduction}

The burgeoning popularity of social media in India has produced enormous amounts of user generated text content. India\rq{}s rich linguistic diversity coupled with its affinity towards English \textemdash{} India has the largest number of speakers of English as a Second Language (ESL) in the world \textemdash{} has led to the online conversations being rife with Code Switching (CS) and Code Mixing (CM). Code Switching is the practice of alternating between two or more languages or varieties of a language in the course of a single utterance \cite{Gumperz:82}. In Code Switching,  unlike Code Mixing where one or more linguistic units of a language such as phrases, words and morphemes are embedded into an utterance of another language \cite{Myers-Scotton:97}, there is a distinct boundary separating the chunks corresponding to each language used in the discourse. So, a combination of language identification and monolingual language taggers could be used for Code Switched utterances. \newcite{Solorio:08} used a Spanish POS tagger and \newcite{Vyas:14} used a Hindi POS tagger in conjunction with English monolingual taggers to handle Spanish-English and Hindi-English code-switched discourses respectively.

Part-of-speech (POS) tagging, the process of assigning each word its proper part of speech, is one of the most fundamental parts of any natural language processing pipeline and it is also an integral part of any syntactic analysis. There are highly accurate monolingual POS taggers available for resource-rich languages like English and French, the state-of-the-art being 97.6\% \cite{Choi:16} and 97.8\% \cite{Denis:09}, in large part due to extensively annotated million word corpora such as PennTreeBank \cite{Santorini:90} and French TreeBank \cite{Abeille:03} respectively. Annotated data for code-mixed data is extremely scarce and the efforts to build a POS tagger for it have mostly advanced through the shared tasks organized at FIRE \cite{Choudhury:14}, EMNLP\cite{Barman:14,Solorio:14} and ICON\cite{Soman:15,Pimpale:16} in the past 2 years. In this paper, we describe our POS tagger for three widely spoken Indian languages (Hindi, Bengali, and Telugu), mixed with English, which was submitted to the shared task organized at ICON 2016. The POS tagger was trained using Conditional Random Fields \cite{Lafferty:01}, which is known to perform particularly well for this task \cite{Toutanova:03} among many other applications in biomedical named entity recognition \cite{Settles:04} and information extraction \cite{Ramesh:16}.
 
\section{Dataset}

The contest task was to predict the POS tags at the word level for code-mixed utterances, collected from WhatsApp, Facebook and Twitter accross three language pairs, English-Hindi (En-Hi), English-Bengali (En-Bn) and English-Telugu (En-Te).

The words were also annotated with certain language tags \textemdash{}  \textit{en} for English, \textit{hi/bn/te} for Hindi, Bengali and Telugu respectively, \textit{univ} for punctuations, emoticons, symbols, @ mentions, hashtags, \textit{mixed} for intra-word language mixing for e.g., \textit{jugaad}ing \footnote{The Hindi noun \textit{jugaad} which means frugal innovation is transformed into an English verb by adding the suffix \textit{ing}.}, \textit{acro} for acronyms like lol, rofl, \textit{ne} for named entities, and \textit{undef} for undefined.

Our submission included models to predict the coarse-grained \cite{Petrov:11} and fine-grained POS tags \cite{Jamatia:15} and was trained in a constrained environment, thus precluding any use of external POS taggers.

\subsection{Code-Mixing Index}
\begin{table}[]
\centering
\begin{tabular}{|l|ll|l|l|}
\hline
\multicolumn{1}{|c|}{\multirow{2}{*}{\textbf{\begin{tabular}[c]{@{}c@{}}Language \\ (English+)\end{tabular}}}} & \multicolumn{2}{c|}{\textbf{CMI}}                                      & \multirow{2}{*}{\textbf{\begin{tabular}[c]{@{}l@{}}Num\\  utt.\end{tabular}}} & \multicolumn{1}{c|}{\multirow{2}{*}{\textbf{\begin{tabular}[c]{@{}c@{}}Mixed\\ (\%)\end{tabular}}}} \\
\multicolumn{1}{|c|}{}                                                                                         & \multicolumn{1}{c}{\textbf{all}} & \multicolumn{1}{c|}{\textbf{mixed}} &                                                                               & \multicolumn{1}{c|}{}                                                                               \\ \hline
Telugu                                                                                                         & 31.94                            & 39.10                               & 989                                                                           & 81.70                                                                                               \\
Hindi                                                                                                          & 11.78                            & 20.06                               & 882                                                                           & 58.73                                                                                               \\
Bengali                                                                                                        & 23.76                            & 24.77                               & 762                                                                           & 95.93                                                                                               \\ \hline
\end{tabular}
\caption{Code-Mixing-Index: Facebook Corpus}
\label{my-label}
\end{table}

\begin{table}[]
\centering
\begin{tabular}{|l|ll|l|l|}
\hline
\multicolumn{1}{|c|}{\multirow{2}{*}{\textbf{\begin{tabular}[c]{@{}c@{}}Language \\ (English+)\end{tabular}}}} & \multicolumn{2}{c|}{\textbf{CMI}}                                      & \multirow{2}{*}{\textbf{\begin{tabular}[c]{@{}l@{}}Num\\  utt.\end{tabular}}} & \multicolumn{1}{c|}{\multirow{2}{*}{\textbf{\begin{tabular}[c]{@{}c@{}}Mixed\\ (\%)\end{tabular}}}} \\
\multicolumn{1}{|c|}{}                                                                                         & \multicolumn{1}{c}{\textbf{all}} & \multicolumn{1}{c|}{\textbf{mixed}} &                                                                               & \multicolumn{1}{c|}{}                                                                               \\ \hline
Telugu                                                                                                         & 34.94                            & 35.37                               & 991                                                                           & 98.79                                                                                               \\
Hindi                                                                                                          & 25.66                            & 28.13                               & 1206                                                                          & 91.21                                                                                               \\
Bengali                                                                                                        & 29.45                            & 29.50                               & 585                                                                           & 99.83                                                                                               \\ \hline
\end{tabular}
\caption{Code-Mixing Index: Twitter Corpus}
\label{my-label}
\end{table}

\begin{table}[]
\centering
\begin{tabular}{|l|ll|l|l|}
\hline
\multicolumn{1}{|c|}{\multirow{2}{*}{\textbf{\begin{tabular}[c]{@{}c@{}}Language \\ (English+)\end{tabular}}}} & \multicolumn{2}{c|}{\textbf{CMI}}                                      & \multirow{2}{*}{\textbf{\begin{tabular}[c]{@{}l@{}}Num\\  utt.\end{tabular}}} & \multicolumn{1}{c|}{\multirow{2}{*}{\textbf{\begin{tabular}[c]{@{}c@{}}Mixed\\ (\%)\end{tabular}}}} \\
\multicolumn{1}{|c|}{}                                                                                         & \multicolumn{1}{c}{\textbf{all}} & \multicolumn{1}{c|}{\textbf{mixed}} &                                                                               & \multicolumn{1}{c|}{}                                                                               \\ \hline
Telugu                                                                                                         & 36.55                            & 36.88                               & 690                                                                           & 99.13                                                                                               \\
Hindi                                                                                                          & 5.88                             & 27.60                               & 981                                                                           & 21.30                                                                                               \\
Bengali                                                                                                        & 0.31                             & 30.05                               & 1052                                                                          & 1.05                                                                                                \\ \hline
\end{tabular}
\caption{Code-Mixing Index: WhatsApp Corpus}
\label{my-label}
\end{table}
\begin{table}[]
\centering
\begin{tabular}{|l|ll|l|l|}
\hline
\multicolumn{1}{|c|}{\multirow{2}{*}{\textbf{\begin{tabular}[c]{@{}c@{}}Language \\ (English+)\end{tabular}}}} & \multicolumn{2}{c|}{\textbf{CMI}}                                      & \multirow{2}{*}{\textbf{\begin{tabular}[c]{@{}l@{}}Num\\  utt.\end{tabular}}} & \multicolumn{1}{c|}{\multirow{2}{*}{\textbf{\begin{tabular}[c]{@{}c@{}}Mixed\\ (\%)\end{tabular}}}} \\
\multicolumn{1}{|c|}{}                                                                                         & \multicolumn{1}{c}{\textbf{all}} & \multicolumn{1}{c|}{\textbf{mixed}} &                                                                               & \multicolumn{1}{c|}{}                                                                               \\ \hline
Telugu                                                                                                         & 11.62                            & 32.60                               & 617                                                                           & 35.66                                                                                               \\
Hindi                                                                                                          & 18.76                            & 23.37                               & 728                                                                           & 80.22                                                                                               \\
Bengali                                                                                                        & 3.71                             & 24.72                               & 3718                                                                          & 15.01                                                                                               \\ \hline
\end{tabular}
\caption{Code-Mixing Index: ICON 2015}
\label{my-label}
\end{table}
In order to compare code-mixed POS taggers trained on different data-sets, it is necessary to have a measure of the code-mixing complexity. Code-Mixing Index(CMI) \cite{Gamback:14} is one such metric that describes the complexity of code-switched corpora and it amounts to finding the most frequent language in the utterance and then counting the frequency of the words belonging to all other languages present. Thus utterances that have only a single language, have a CMI of 0. 

Tables 1, 2, 3, and 4, show the following CMI metrics that were calculated for Facebook, Twitter, WhatsApp data of 2016 and the training data of ICON 2015 respectively.
\begin{enumerate}
\item \textit{CMI all:} average CMI for all sentences in a corpus
\item \textit{CMI mixed:} average CMI for the sentences with non-zero CMI.
\item \textit{Mixed \%:} percentage of code-mixed sentences in the corpus
\item \textit{Num utt.:} total number of utterances in the corpus.
\end{enumerate}
We observed that the WhatsApp corpus for Bengali has a very low fraction of code-mixed sentences i.e., there are an extremely low number of words tagged as \textit{en} in the data-set. On closer inspection of the dataset, there were exactly 13 instances of words that were tagged \textit{en} and these were actually words such as \textit{Kolkata} and \textit{San Antonio}, that should have been annotated as \textit{ne} instead. Effectively, CMI for WhatsApp-Bengali corpus is 0.

\section{Model and Results}
POS tagging is considered to be a sequence labelling task, where each token of the sentence needs to be assigned a label. These labels are usually interdependent, because the sentence follows grammar rules inherent to the language.

We have used the CRF implementation of \textit{sklearn-crfuite}\footnote{http://sklearn-crfsuite.readthedocs.io/en/latest} because it is particularly well suited for sequence labelling tasks.   

\subsection{Features}
The feature-set consisted of character-case information, character n-grams of gram size upto 3, which would thereby also encompass all prefixes and suffixes, patterns for email and web-site urls, punctuations, emoticons, numbers, social media specific characters like @,\# and also the language tag information.

We chose a CRF window size of two and performed grid-search to choose the best optimization algorithm and L1/L2 regularization parameters\footnote{Our code is available at https://github.com/le-scientifique/code-mixing-social-media}. There were a total of 18 models trained using this pipeline, one for each case in the cross-product:

\centerline{\{bn-en, hi-en, te-en\}}
\centerline{X} 
\centerline{\{WhatsApp, Twitter, Facebook\}}
\centerline{X} 
\centerline{\{Fine-Grained, Coarse-Grained\}}

\subsection{Results}
The F1 measure of our model against the social networks is depicted in Table 5 and the results with respect to the POS granularity is shown in Table 6. These results were calculated on the private test data-set shared by the organizers. With the system described in the paper, we achieved an overall average score of 76.45\%, across all 18 models. This is only marginally lesser than 76.79\%, which was the the score of winning entry of ICON 2015, and we are awaiting the results of ICON 2016. 

\begin{table}[]
\centering
\begin{tabular}{|l|l|l|l|}
\hline
\textbf{\begin{tabular}[c]{@{}l@{}}Language\\ (English +)\end{tabular}} & \textbf{WhatsApp} & \textbf{Twitter} & \textbf{Facebook} \\ \hline
\textbf{Telugu}                                                         & 74.43             & 79.15            & 74.10             \\
\textbf{Hindi}                                                          & 75.68             & 86.80            & 77.44             \\
\textbf{Bengali}                                                        & 76.71             & 69.64            & 74.1              \\ \hline
\end{tabular}
\caption{Model Performance (F1-Score) w.r.t Social Networks}
\label{my-label}
\end{table}

\begin{table}[]
\centering
\begin{tabular}{|l|l|l|}
\hline
\textbf{\begin{tabular}[c]{@{}l@{}}Language\\ (English +)\end{tabular}} & \textbf{\begin{tabular}[c]{@{}l@{}}Fine-\\ Grained\end{tabular}} & \textbf{\begin{tabular}[c]{@{}l@{}}Coarse-\\ Grained\end{tabular}} \\ \hline
\textbf{Telugu}                                                         & 73.50                                                            & 78.30                                                              \\
\textbf{Hindi}                                                          & 83.40                                                            & 76.60                                                              \\
\textbf{Bengali}                                                        & 73.28                                                            & 76.39                                                              \\ \hline
\end{tabular}
\caption{Model Performance (F1-Score) w.r.t POS Granularity}
\label{my-label}
\end{table}

\section{Conclusion \& Future Work}
In this paper, we presented a CRF based POS tagger for code-mixed social media text in the constrained environment, without making use of any external corpora or monolingual POS taggers. We achieved an overall F1- Score of 76.45\%.  We would like to evaluate the performance improvement or lack thereof upon training a POS tagger in an unconstrained environment by utilizing monolingual taggers trained on Indic languages. Multilingual tools are still a ways off from matching the state-of-the-art of the tools available for monolingual linguistic analysis. There is promising research in the field of developing tools for resource poor languages by applying Transfer Learning \cite{Zoph:16}, which could also be evaluated in the future. Upon inspecting the dataset, we observed a few inaccuracies in annotation, which could be addressed by leveraging crowd-sourcing platforms that can execute Human Intelligence Tasks.

\end{document}